\documentclass[letterpaper, 10 pt, conference]{ieeeconf}  

\IEEEoverridecommandlockouts                              

\usepackage{cite}
\usepackage{amsmath,amssymb,amsfonts}
\usepackage{algorithmic}
\usepackage{graphicx}
\usepackage{textcomp}
\usepackage{xcolor}

\usepackage{pifont}
\newcommand{\cmark}{\ding{51}}

\usepackage{multirow}
\usepackage{booktabs}
\usepackage{tabularx}
\usepackage{bm}
\usepackage{amsmath}
\usepackage{dsfont}
\usepackage{amssymb}
\usepackage{physics}
\usepackage[pagebackref=true,breaklinks=true,colorlinks,bookmarks=false]{hyperref}

\title{\LARGE \bf
Exploring Visual Pre-training for Robot Manipulation: Datasets, Models and Methods
}

\author{Ya Jing$^{1,*}$, Xuelin Zhu$^{1,2,*}$, Xingbin Liu$^{1,*}$, Qie Sima$^{1,3}$, Taozheng Yang$^{1}$, Yunhai Feng$^{1}$, Tao Kong$^{1\ddagger}$ \\
$^1$ByteDance Research, ~$^2$Southeast University, ~$^3$Tsinghua University \\
\url{https://explore-pretrain-robot.github.io}
\thanks{*Equal contribution.}
\thanks{$^\ddagger$Corresponding author: Tao Kong (\texttt{kongtao@bytedance.com}).} 
}%

\begin{document}

\maketitle
\thispagestyle{empty}
\pagestyle{empty}

\begin{abstract}
Visual pre-training with large-scale real-world data has made great progress in recent years, showing great potential in robot learning with pixel observations. 
However, the recipes of visual pre-training for robot manipulation tasks are yet to be built.
In this paper, we thoroughly investigate the effects of visual pre-training strategies on robot manipulation tasks from three fundamental perspectives: pre-training datasets, model architectures and training methods.
Several significant experimental findings are provided that are beneficial for robot learning. 
Further, we propose a visual pre-training scheme for robot manipulation termed Vi-PRoM, which combines self-supervised learning and supervised learning. 
Concretely, the former employs contrastive learning to acquire underlying patterns from large-scale unlabeled data, while the latter aims learning visual semantics and temporal dynamics. 
Extensive experiments on robot manipulations in various simulation environments and the real robot demonstrate the superiority of the proposed scheme. 
Videos and more details can be found on \url{https://explore-pretrain-robot.github.io}.

\end{abstract}


\section{INTRODUCTION}
\label{sec:intro}
The past years have witnessed substantial progress in visual representation learning based on deep neural networks. 
After pre-training on large-scale visual data, the neural network is subsequently employed as a general-purpose encoder to extract visual representations for many tasks, e.g., image segmentation~\cite{long2015fully}, object detection \cite{ren2015faster} and autonomous driving \cite{zhang2022action}, showing its strong generalization ability, while also highlighting its potential in robot manipulation. 

Learning from visual observations for robot manipulation is known as a challenging task that requires a thorough understanding of both visual semantics and sequential patterns of observations. A common method is to train the visual encoder and model-based policy from scratch in an end-to-end manner with in-domain data \cite{levine2016end,kalashnikov2018scalable}. 
Despite its effectiveness to some degree, such a method requires training on a large number of observation-action samples, which may limit its wide applications.
Therefore, pre-training the visual encoder with large-scale off-the-shelf data from the real world can serve as an alternative. Benefiting from its strong generalization ability, the pre-trained visual encoder is expected to generalize across a range of robot manipulation tasks and enable data-efficient learning. 

\begin{figure}[t]
\centering
\includegraphics[width=0.83\linewidth]{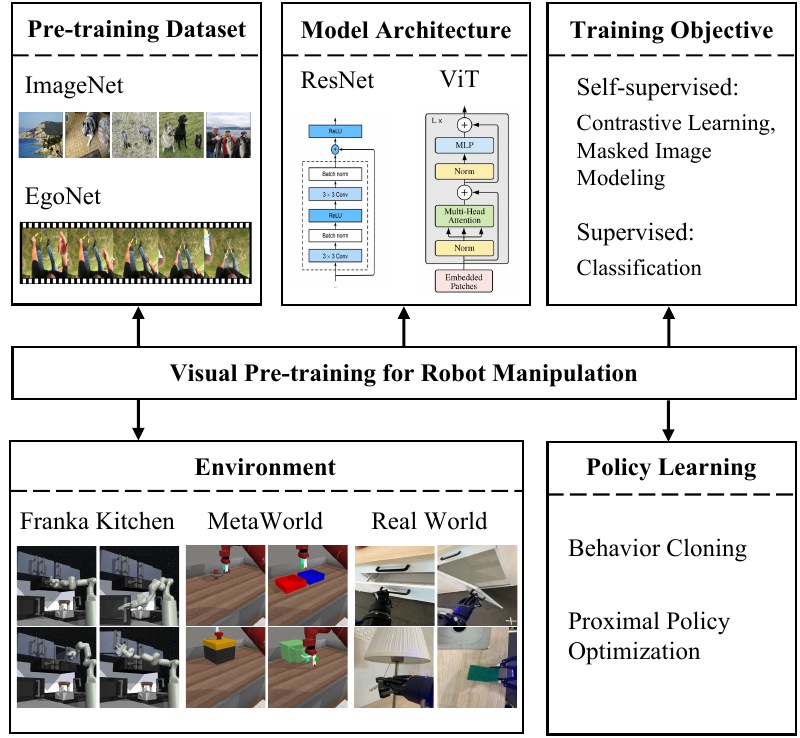}
\caption{General path of visual pre-training for robot manipulation.}
\label{fig:illu}
\end{figure}

Recently, visual pre-training on large-scale real-world data for robot learning has attracted increasing interest. Prominent performance gains reported on prior works \cite{nair2022r3m,mvp} show its great potential in learning robot control from pixels. Despite the claimed advantage, these works differ in pre-training data, methods and models. So it remains an open question about which types of data, pre-training methods and models can better assist robot manipulation. A system-level benchmark on the profits of visual pre-training is in demand. 

In this paper, as shown in Figure \ref{fig:illu}, we first conduct extensive studies on visual pre-training from three fundamental aspects: datasets, models and methods that may influence the performance of robot learning. 
Hopefully, these can facilitate future research in the community.
Based on empirical findings, we propose a visual pre-training scheme oriented for robot manipulations, which sequentially trains a visual encoder using self-supervised learning and supervised fine-tuning. Concretely, the visual encoder is first pre-trained based on contrastive learning~\cite{mocov3}, allowing the trained model to acquire sequential patterns implicitly for the input data. 
Then, supervised learning is applied by constructing pseudo-labels and temporal labels to encourage the visual encoder further to perceive visual semantics and temporal dynamics. 
In addition, we propose a new dataset named EgoNet, which is created based on Ego4d \cite{grauman2022ego4d} and contains a large-scale egocentric video clips rich in human-object interactions. EgoNet has the potential to serve as a benchmark to pre-train visual models for robot manipulations. 

Our main contributions are summarized as: (1) We create the EgoNet dataset, a new benchmark enriched with diverse scenarios and human-object interactions for robotic visual pre-training.  (2) We fully explore the visual pre-training in terms of datasets, methods and models, and provide several key suggestions 
for robot manipulation tasks. (3) We propose a novel cascade visual pre-training scheme that enables the visual encoder to learn sequential patterns, visual semantics and temporal dynamics from the large-scale real-world data, and achieves remarkable performance improvement on robot manipulation tasks.

\section{RELATED WORK}
\label{sec:related}
\subsection{Vision-Based Robot Learning}
The robotic community has long focused on vision-based learning methods for various robot tasks in the past decade. Currently, the most prevailing paradigm of vision-based robot learning is the end-to-end method \cite{kalashnikov2018scalable}. With the surge of deep learning in the last decade, many CNN-based models have been proposed to enable the visual modality of robots in manipulation tasks \cite{redmon2015real,kumra2017robotic}. Furthermore, CNN-RNN methods \cite{shridhar2020ingress,Zhang2021INVIGORATE}
are widely adopted to solve the task of human instruction in natural language. 
Recently, many methods \cite{nair2022r3m,mvp,maskvit} based on pre-trained models have been proposed for robot learning. 
Several previous methods investigated the self-supervised pre-training in robot manipulation, e.g., R3M~\cite{nair2022r3m}, MVP~\cite{mvp}, and MaskViT~\cite{maskvit}. 
These works focus on one side of visual pre-training, thus calling for a systematic study.

\subsection{Representation Learning}
Self-supervised visual pre-training has been an active research topic recently, and can learn universal visual representations. 
Visual pre-training aims to learn visual representations by masked image modeling~\cite{mae,videomae} and contrastive learning~\cite{moco, mocov3}.
While the vision-language pre-training aims to
learn the semantic correspondence between different modalities \cite{chen2020uniter,li2021align}.
Pre-training datasets are significant for the representation learning.
To learn reusable representations that can generalize well to robotic manipulation tasks, the interaction between humans and objects needs to be captured. Recently, a diverse and large-size dataset Ego4D \cite{grauman2022ego4d} has been proposed, which contains daily-life activity videos spanning hundreds of scenarios.

\subsection{Robot Manipulation Benchmarks}
With the recent progress in exploiting the pre-trained models in robotic tasks, a number of robotic manipulation benchmarks have been introduced to evaluate the performance of the pre-trained model. The off-shelf robotic manipulation benchmarks can be categorized into two main kinds by simulators: RL (Reinforcement Learning) benchmarks and embodied benchmarks. The RL benchmarks focus on the training and evaluation of reinforcement learning agents where a simulated environment with several robot models and scenarios in limited space is usually provided. 
Recent RL benchmarks explore the training and evaluation of robotic manipulation method in aspects of multi-task training \cite{yu2020meta}, more realistic scenarios with clutter \cite{gupta2020relay}, tasks in higher complex level \cite{rajeswaran2017learning}, more kinds of manipulation forms \cite{RLBench} and manipulations with linguistic instructions\cite{zheng2022vlmbench,mees2022calvin}. Meanwhile, the pre-trained models are widely introduced as solutions to robot manipulation tasks.

\label{sec:method}
\section{Benchmarking}

\begin{figure}[t]
\centering
\includegraphics[width=0.95\linewidth]{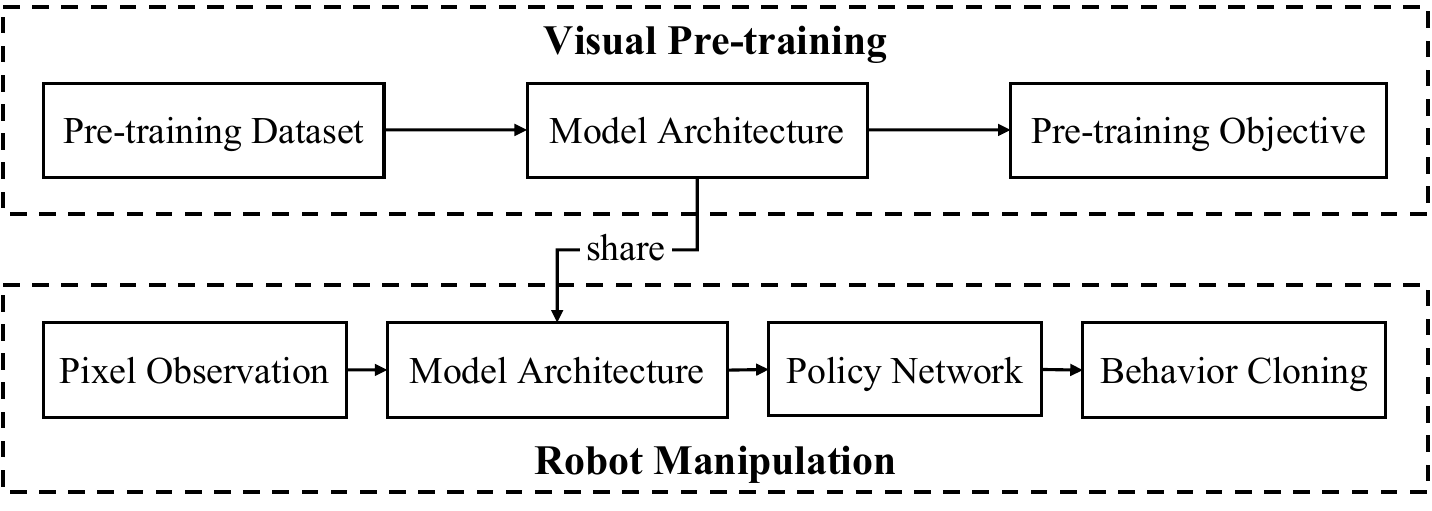}
\caption{The study pipeline of visual pre-training for robot manipulation.}
\label{fig:pipeline}
\end{figure}

In this section, we explore key components that affect the pre-training behaviors and the robot manipulation performance, i.e., pre-training datasets, optimization methods, and model architectures.
The study pipeline is shown in Figure~\ref{fig:pipeline}. We first pre-train the visual encoder on the pre-training dataset.
Then we adopt typical imitation learning methods on robot manipulation tasks to verify the effectiveness of visual representations, where the encoder parameters are frozen during training. 
In this way, we could give system-level studies of each component.

\subsection{Benchmarking Setup}

To evaluate the effectiveness of the pre-trained visual encoder, we adopt two robot control simulation environments, i.e., Franka Kitchen \cite{gupta2019relay} and MetaWorld \cite{yu2020meta}, for robot learning. As shown in the right part of Figure \ref{fig:model}, we choose the same tasks as \cite{nair2022r3m}.
Please refer to Section \ref{sec:exp} for the pre-training details and evaluation metrics.

\subsubsection{Pre-training Dataset}
ImageNet \cite{deng2009imagenet} has recently been widely used in self-supervised pre-training for various downstream tasks.
However, ImageNet lacks dynamic interaction between objects, making it may be unsuitable to serve as pre-training data for robot manipulation tasks. 

We propose a new benchmark, called EgoNet, to pre-train visual encoders for robot manipulation. It comprises nearly 500,000 video clips covering hundreds of scenarios and is rich in human-object interactions. The EgoNet is constructed based on Ego4D \cite{grauman2022ego4d}.
We experimentally intercept a short clip with a duration of 1\textit{s} for each narration. With this strategy, a total of
0.503 million video clips rich in human-object interactions are collected. Note that the video in Ego4D has a frame rate of 30 fps. After a 10-fold uniform down-sampling, EgoNet is obtained that contains about 1.5 million video frames in total, making the training samples number comparable with ImageNet.

\subsubsection{Model Architecture}
The architecture of visual encoder is also an important element in determining the performance of robot manipulation tasks. To explore its effect, we choose three typical models, namely convolution-based ResNet-34~\cite{he2016deep}, ResNet-50~\cite{he2016deep}, and ResNet-101~\cite{he2016deep}, 
which have been the defacto standard for visual representation extraction. In this way, we could provide insight into which architectures are more beneficial for robot manipulation tasks.

\subsubsection{Pre-training Method}
The learning objective directly determines the type of representations that the model can learn from a dataset. Contrastive learning and masked image modeling, the two most prevalent pre-training methods in self-supervised learning, are naturally the main exploration goals in this work. 
Contrastive learning aims to encourage the feature similarity between two different augmented views of the same image but suppress the similarity between different images.  Masked image modeling resorts to reconstructing the randomly masked patches of the input image. 
In this work, we choose MoCo-v3~\cite{mocov3} and MAE (Masked AutoEncoder)~\cite{mae} for contrastive learning and masked image modeling, respectively. 

\subsection{Main Observations}
\subsubsection{Pre-training Dataset}
\textbf{EgoNet is more powerful than ImageNet.}
We pre-train visual encoder (i.e., ResNet-50)
on different datasets, i.e., ImageNet and EgoNet, using the contrastive learning method (MoCo-v3), and observe their performance on the robot manipulation tasks. From Table \ref{tab:data}, we can see that the model pre-trained on EgoNet 
achieve better performance on robot manipulation tasks. Obviously, the robot favors the interaction-related knowledge and temporal relationships contained in the video in terms of manipulation tasks. In addition, the egocentric natural images in EgoNet have much more global context about the world, which means richer visual features can be learned. 

\newcolumntype{Z}{p{1.5cm}<{\centering}}
\begin{table}[t]
\centering
\caption{Effects of pre-training datasets on robot manipulation on two simulators, i.e., Franka Kitchen and MetaWorld, using success rate (\%) as the metric.\label{tab:data}}
\begin{tabularx}{\linewidth}{Z|Z|X<{\centering}|X<{\centering}}
\toprule
Model & Dataset & {Franka Kitchen} & {MetaWorld} \\
\hline
\hline
ResNet-50 & ImageNet & 31.1 & 54.1 \\
\hline
ResNet-50 & EgoNet & \textbf{40.5} & \textbf{61.2} \\
\bottomrule
\end{tabularx}
\end{table}

\newcolumntype{Z}{p{1.5cm}<{\centering}}
\begin{table}[t]
\centering
\caption{Effects of model architectures on robot manipulation. \label{tab:model}}
\begin{tabularx}{\linewidth}{Z|Z|X<{\centering}|X<{\centering}}
\toprule
Model & Dataset & {Franka Kitchen} & {MetaWorld} \\
\hline
\hline
ResNet-34 & EgoNet & 22.6 & 52.4 \\
\hline
ResNet-50 & EgoNet & \textbf{40.5} & 61.2 \\
\hline
ResNet-101 & EgoNet & 40.0 & \textbf{61.6} \\
\bottomrule
\end{tabularx}
\end{table}

\subsubsection{Model Architecture}
\textbf{ResNet-50 performs better.}
From Table \ref{tab:model}, we can observe that ResNet-50 and ResNet-101 perform better than ResNet-34 on the robot manipulation tasks in both simulation environments
pre-trained on EgoNet. In addition, there is no performance improvement as the model increases from ResNet-50 to ResNet-101. Furthermore, recent work suggests that pre-training ViT \cite{dosovitskiy2020image} models with larger pre-trained datasets can achieve better results. 

\begin{figure*}[t]
\centering
\includegraphics[width=\linewidth]{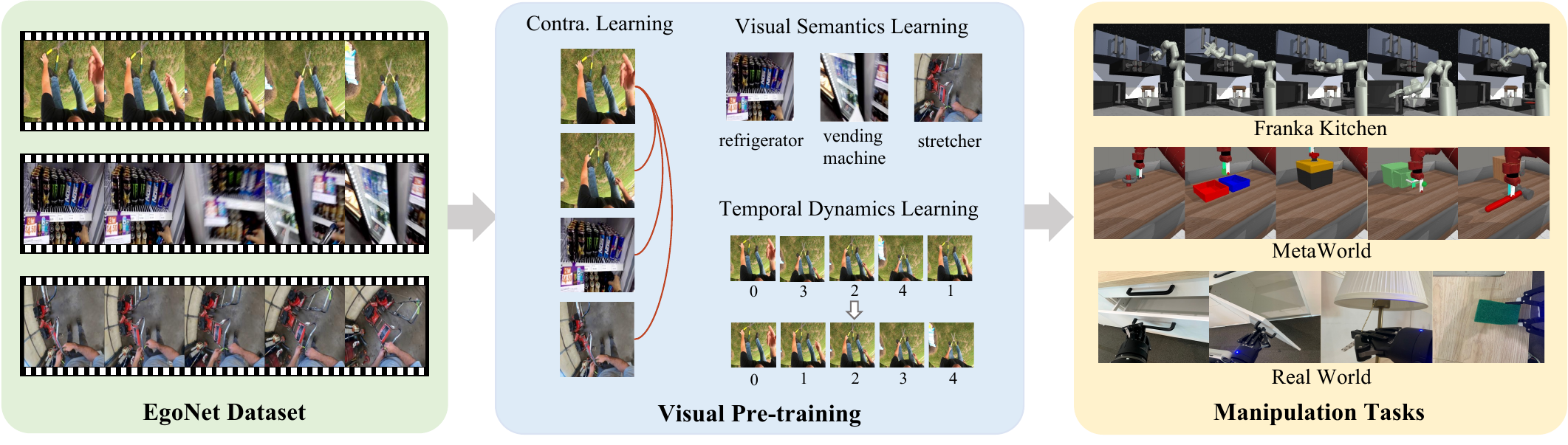}
\caption{The pipeline of our Vi-PRoM. The EgoNet dataset is first constructed to serve as pre-training data. We first pre-train the ResNet-50 with contrastive learning, enabling the model to learn universal visual representations. Then, frame-order and pseudo-label predicting tasks are jointly applied to encourage the model to capture temporal and semantic visual representations. Note that the pseudo-labels are automatically generated by the ResNet-101 model pre-trained on ImageNet without manual labeling. Finally, the pre-trained model is utilized to extract visual representations for robot manipulation tasks.}
\label{fig:model}
\end{figure*}

\subsubsection{Pre-training Method}
\textbf{Contrastive learning is preferred.}

As shown in Table \ref{tab:method}, MoCo-v3 outperforms MAE on both ImageNet and EgoNet datasets, demonstrating the effectiveness of contrastive learning compared to masked image modeling for manipulation. This result also suggests that the visual semantics acquired by contrastive learning are more important for robot manipulation than the structural information learned by masked image modeling.

\newcolumntype{Z}{p{1.8cm}<{\centering}}
\begin{table}[t]
\centering
\caption{Effects of pre-training methods on robot manipulation. 
\label{tab:method}}
\begin{tabularx}{\linewidth}{X<{\centering}|p{1.5cm}<{\centering}|Z|p{1.5cm}<{\centering}}
\toprule
Learning Method & Dataset & {Franka Kitchen} & {MetaWorld} \\
\hline
\hline
MAE~\cite{mae} & ImageNet & 11.4 & 50.3 \\
\hline
MoCo-v3~\cite{mocov3} & ImageNet & \textbf{31.1} & \textbf{54.1} \\
\hline
\hline
MAE~\cite{mae} & EgoNet & 18.0 & 49.8 \\
\hline
MoCo-v3~\cite{mocov3} & EgoNet & \textbf{40.5} & \textbf{61.2} \\
\bottomrule
\end{tabularx}
\end{table}

\subsection{Summary}
Through the aforementioned explorations on various pre-training datasets, model architectures and pre-training methods, three key conclusions could be drawn: 
\begin{itemize}
    \item Visual pre-training with human-object interaction data is of great importance for robot manipulation.
    \item Convolution-based ResNet-50 is preferred in retaining visual knowledge for robot manipulation.
    \item The sequential pattern and semantic information learned by contrastive learning are more effective.
\end{itemize}

\section{Proposed Approach}
Based on the above explorations, we propose \underline{Vi}sual \underline{P}re-training scheme for \underline{Ro}bot \underline{M}anipulation (Vi-PRoM), which pre-trains ResNet-50 on the EgoNet dataset to extract comprehensive visual representations for robot manipulation. Specifically, we first employ contrastive learning to acquire human-object interaction patterns from the EgoNet dataset in a self-supervised manner. Then two additional learning objectives, i.e., visual semantics predicting and temporal dynamics predicting, are proposed to further enrich the encoder. Figure \ref{fig:model} shows a basic pipeline of the proposed Vi-PRoM. Note that we do not need manually annotate the labels to learn both visual semantics and temporal dynamics.

\subsection{Contrastive Self-supervised Learning}
We hypothesize a good visual representation should have the ability to distinguish different scenes.
Therefore, we use contrastive learning as our self-supervised paradigm to let the model learn rich and general visual representations.
The contrastive objective function pulls features generated by similar images together and pushes the features generated by different images away. Specifically, we sample a minibatch of images and minimize the InfoNCE loss~\cite{simclr}.

\subsection{Supervised Learning} 
With the learned representation from contrastive learning, it is imperative to learn visual semantics and temporal dynamics to generalize well for robot manipulation.

\subsubsection{Learning Visual Semantics}
We introduce a pseudo-label predicting task to fine-tune the learned backbone, encouraging the model to learn better visual semantic representations. 
Specifically, we employ a ResNet-101 model supervised on ImageNet to generate pseudo labels for EgoNet. Then, the pseudo label is used to fine-tune our self-supervised learned backbone with the cross-entropy loss:
\begin{equation}
    \mathcal{L}_\text{VS} = - \mathbb E_D \sum_{i=1}^{N} \mathcal{T}(x_i) \log(h_1(f(x_i))),
\end{equation}
where $D$ is the EgoNet dataset, $\mathcal{T}$ is the ResNet-101 network to generate pseudo labels for each sample $x_i$, $f$ is the backbone, and $h_1$ is a classification head.

\subsubsection{Predicting Temporal Dynamics}
Robot manipulation tasks require predicting the next actions based on current and historical observations. Thus they are sensitive to temporal dynamics. We design a frame order prediction task to enable the model to learn temporal dynamics for each clip of EgoNet. Given the image set $\mathcal{I} = \{x_0,...,x_k,...,x_{N-1} \}$ sampled sequentially from a video clip, we scramble these images and then predict the original order for the image $x_k$. This task is formulated as a classification problem of $N$ classes, which is commonly solved by minimizing the cross-entropy loss:

\begin{equation}
  \mathcal{L}_\text{TD} = - \mathbb E_D \sum_{k=0}^{N-1}\mathbf{y}_{k}\log(h_2(f(x_k))),
\end{equation}
where $h_2$ is a classification head. $\mathbf{y}_k$ denotes the order of the image $x_k$ in original image set $\mathcal{I}$.

\subsubsection{Joint Training}
We combine the visual semantics and temporal dynamics loss for jointly training:
\begin{equation}
    \mathcal{L}_\text{total} = \mathcal{L}_\text{VS} + \lambda \mathcal{L}_\text{TD},
\end{equation}
where $\lambda$ is the balance coefficient set as 0.33 in practice.
In principle, visual semantics and temporal dynamics predicting together guide the learning, enabling the model to learn semantic and temporal visual representations.

\subsection{Robot Imitation Learning}
Given the well-trained visual encoder $f$, the robot utilizes it to encode visual features of pixel observations for policy learning. In this work, we employ the typical behavior cloning (BC) \cite{torabi2018behavioral} method to imitate expert demonstrations, where the policy network is parameterized as a two-layer perceptron. 

\section{EXPERIMENTS}
\label{sec:exp}
\subsection{Experimental Setup \label{sec:exp}}

To evaluate the pre-trained visual encoder on robot manipulation tasks, we take it as a frozen module for policy learning. We train the policy network for 20,000 steps using a batch size of 32 and an Adam optimizer with a learning rate of 0.001. Unless otherwise specified, the demonstration dataset size used for imitation learning is set as 5. In the PPO experiments, we train for 20 iterations with 10 epochs per iteration. The reward function we use is similar to \cite{xiao2022masked}. The average of the best success rates on all manipulation tasks with three different seeds (100, 125, 150) is reported to measure the performance of the visual encoder. 

In the real environment, our robot hardware is mainly formed by a differential-drive mobile base equipped with a 2d LiDAR and IMU and a 6-DoF arm. A 2-finger parallel gripper is equipped for contact-rich interactions. Between the end-effector and the arm, a force torque sensor is installed to measure the forces and torques experienced by the robot, which is utilized to stop the robot if any large forces or torques appear. The robot's wrist is equipped with an RGBD camera as its perception unit. The Intel core i7 CPU is chosen as the computing unit.

\newcolumntype{Z}{p{1.25cm}<{\centering}}
\begin{table}[t]
\centering
\caption{Comparison results with the state-of-the-art methods.  \label{tab:sota}}
\begin{tabularx}{\linewidth}{l|Z|Z|Z|Z}
\toprule
\multirow{2}{*}{Method} & \multicolumn{2}{c|}{Franka Kitchen} & \multicolumn{2}{c}{MetaWorld} \\
\cline{2-5}
 & BC & PPO & BC & PPO \\
\hline
\hline
Scratch & 22.3 & 15.2 & 26.5 & 28.8 \\
\hline
R3M \cite{nair2022r3m} & 27.4 & 18.3 & 61.7 & 38.6 \\
\hline
MoCo-v3 \cite{mocov3} & 40.5 & 36.8 & 61.2 & 43.6 \\
\hline
\textbf{Vi-PRoM} & \textbf{43.8} & \textbf{39.5} & \textbf{63.5} & \textbf{46.8} \\
\bottomrule
\end{tabularx}
\end{table}

\newcolumntype{Z}{p{1.4cm}<{\centering}}
\begin{table}[h]
\centering
\caption{Ablation study on different modules. \label{tab:abl}}
\begin{tabularx}{\linewidth}{Z|Z|Z|X<{\centering}|X<{\centering}}
\toprule
Contrastive Learning & Visual Semantics & Temporal Dynamics & Franka Kitchen & MetaWorld \\
\hline
\hline
\cmark &  &   & 40.5 & 61.2 \\
\hline
\cmark & \cmark &   & 43.2 & 62.0 \\
\hline
\cmark &  & \cmark & 40.7 & 62.6 \\
\hline
\cmark & \cmark & \cmark & \bf{43.8} & \bf{63.5} \\
\bottomrule
\end{tabularx}
\end{table}

\subsection{Main Results}
\subsubsection{Simulation Environments}
To demonstrate the effectiveness of our Vi-PRoM, we compare it with the state-of-the-art visual pre-training methods for robot manipulation. For fair comparisons, except for the scratch method, whose visual encoder parameters are randomly initialized, all other models are pre-trained on our EgoNet dataset and evaluated with the behavior cloning method. Note that the visual encoder for each method is ResNet-50. Experimental results are reported in Table~\ref{tab:sota}. 
It can be seen that our model achieves the best performance in both simulation environments. 
In addition, the performance gains of our Vi-PRoM over the MoCo-v3, reaching 3.3\% and 2.3\% in success rate in Franka Kitchen and MetaWorld, respectively, indicate the value of explicitly learning visual semantics and temporal dynamics.

To learn the temporal dynamics and visual semantics, R3M resorts to the time contrastive learning and video-language alignment. Compared with R3M, our Vi-PRoM shows considerable performance gains, especially in the Franka Kitchen environment.
Notably, in terms of the capacity to learn visual semantics and temporal dynamics, our pseudo-label predicting and frame order modeling outperform the time contrastive learning and video-language alignment. 

To further verify the effectiveness of our Vi-PRoM,
we choose the proximal policy optimization (PPO) algorithm \cite{vinyals2015show} as an alternative to behavior cloning.
Experimental results are provided in Table \ref{tab:sota}. Our Vi-PRoM consistently outperforms all competitors on both learning algorithms. 

\subsubsection{Real Robot}
\begin{figure}[t]
	\centering
	\includegraphics[width=0.9\linewidth]{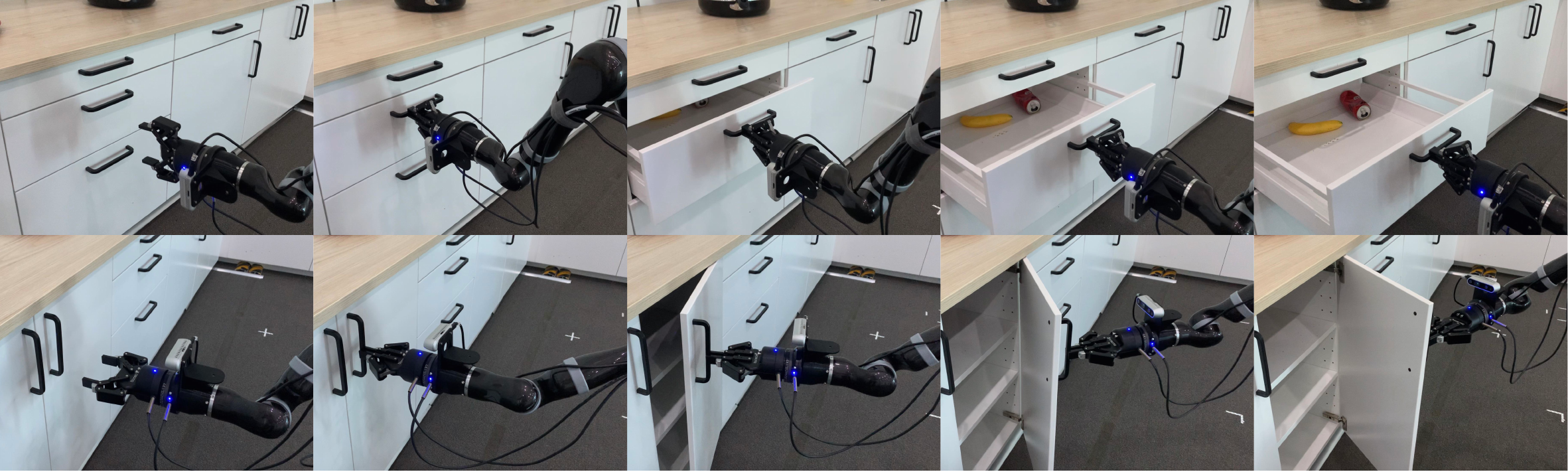}
	\caption{The real robot is able to successfully open the drawer and the door with the help of our Vi-PRoM model in a kitchen environment. \label{fig:real}}
\end{figure}

We deploy our model on a real robot to demonstrate its performance in the real environment. In practice, we test our pre-trained representations on four tasks, i.e., \texttt{opening the door}, \texttt{closing the door}, \texttt{opening the drawer} and \texttt{closing the drawer}. 
We collect 30 demonstrations for each task. Figure \ref{fig:real} shows two successful cases of our model in the real robot environment.
Overall, benefiting from the powerful representational capability of Vi-PRoM, the robot is competent for various manipulation tasks in the real kitchen environment by learning from demonstrations. 

\begin{figure}[t]
	\centering
	\includegraphics[width=0.95\linewidth]{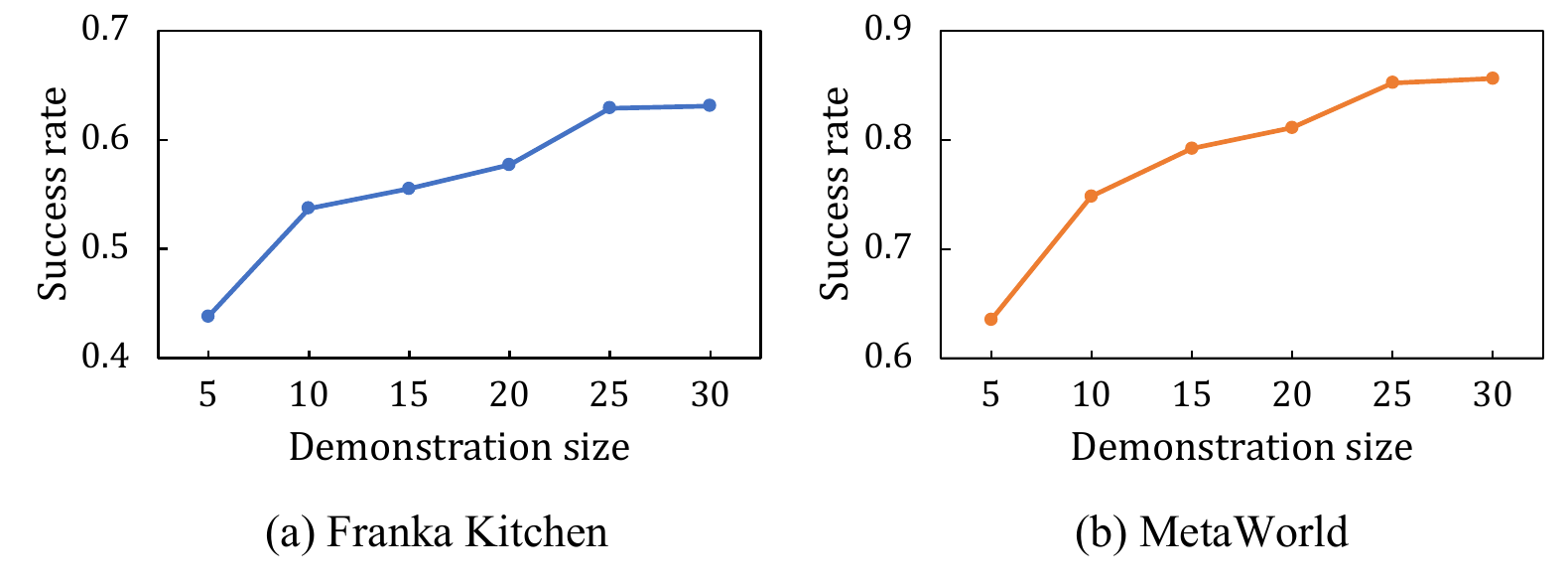}
	\caption{Effects of demonstration size on robot manipulation tasks.\label{fig:demos}}
\end{figure}

\newcolumntype{Z}{p{2.3cm}<{\centering}}
\begin{table}[t]
\centering
\caption{Performance of official models on robot manipulation tasks.\label{tab:cur_model}}
\begin{tabularx}{\linewidth}{l|Z|Z}
\toprule
{Method} & {Franka Kitchen} & {MetaWorld} \\
\hline
\hline
ImageNet Supervised \cite{deng2009imagenet} & 16.5 & 51.9 \\
\hline
MDETR \cite{kamath2021mdetr} & 19.8 & 59.5 \\
\hline
CLIP \cite{radford2021learning} & 11.0 & 53.0 \\
\hline
MAE \cite{mae} & 11.4 & 50.3 \\
\hline
MVP \cite{mvp} & 11.4 & 53.2 \\
\hline
MoCo-v3 \cite{moco} & 31.1 & 54.1 \\
\hline
Vi-PRoM & \textbf{43.8}& \textbf{63.5} \\
\bottomrule
\end{tabularx}
\end{table}

\subsection{Ablation Study}
\subsubsection{Pre-training Components}
Table~\ref{tab:abl} exhibits the experimental results. When visual semantic learning is absent, the success rate decreases by 3.1\% and 0.9\% on Franka Kitchen and MetaWorld, respectively. Analogously, a drop in success rate of 0.6\% and 1.5\% on Franka Kitchen and MetaWorld can be observed in the absence of temporal dynamics learning. These two experimental results demonstrate the importance of visual semantics learning and temporal dynamics learning for robot manipulation. Moreover, when both learning objectives are absent, the success rate of Vi-PRoM suffers from considerable performance degradation. Therefore, the effectiveness of the collaboration between visual semantic learning and temporal dynamics learning is proved.

\subsubsection{Model Scalability}
We also investigate the scalability of Vi-PRoM. As shown in Figure~\ref{fig:demos}, in both the Franka Kitchen and MetaWorld simulation environments, the success rate of Vi-PRoM improves steadily as the size of the demonstration data increases. 
After training on the larger expert demonstration dataset, our proposed Vi-PRoM model shows its scalability on robot manipulation tasks.

\subsubsection{Other Models}
We also report experimental results of directly taking the popular pre-trained models as visual encoders for robot manipulation, as shown in Table ~\ref{tab:cur_model}. ImageNet Supervised \cite{deng2009imagenet} is the ResNet-50 pre-trained for ImageNet classification task. MDETR \cite{kamath2021mdetr} is the ResNet-101 pre-trained on large-scale image-text pairs. CLIP \cite{radford2021learning} is the ResNet-50 trained to align the image representation with the paired text. MAE is the ViT-Base trained on ImageNet. MVP is the ViT-large trained on Ego4D. It can be seen that all these models largely lag behind our Vi-PRoM model.

\section{DISCUSSION AND LIMITATION}
\label{sec:conclusion}
In this paper, we have explored three crucial components that affect the pre-trained model on robot manipulation tasks. Key conclusions are drawn that robot manipulation prefers human-object interaction dataset, convolution-based ResNet-50 network, as well as temporal and semantic information. We further propose the Vi-PRoM for robot manipulation. 
Extensive experiments on simulators and the real environment demonstrate its superiority.

Although our pipeline is effective, there are still many issues to be further explored. First, training visual encoders directly on video clips has the potential to learn better temporal dynamics.
Then using larger pre-training datasets is also worth exploring in the future.
Finally, currently visual encoders are pre-trained on real-world data but evaluated in simulation environments. The significant gap can lead to some unexpected results, and also inspires us to consider establishing an evaluation benchmark from the real environment to facilitate research. 

\bibliographystyle{IEEEtran}
\bibliography{refer}

\end{document}